\documentclass[conference]{IEEEtran}
\IEEEoverridecommandlockouts

\def\BibTeX{{\rm B\kern-.05em{\sc i\kern-.025em b}\kern-.08em
    T\kern-.1667em\lower.7ex\hbox{E}\kern-.125emX}}
\usepackage{balance}
\usepackage{resizegather}
\usepackage{algorithm}
\usepackage{array}
\usepackage[caption=false,font=normalsize,labelfont=sf,textfont=sf]{subfig}
\usepackage{algpseudocode}
\usepackage{textcomp}
\usepackage{csquotes}
\usepackage{stfloats}
\usepackage{tabstackengine}
\usepackage{multirow}  
\usepackage{booktabs}  
\usepackage{threeparttable}  
\usepackage{url}
\usepackage{verbatim}
\usepackage{graphicx}
\usepackage{cite}
\usepackage{amssymb,amsmath,amsthm,enumitem}

\usepackage[cmintegrals]{newtxmath}
\usepackage[export]{adjustbox} 
\usepackage[acronym]{glossaries}
\newacronym{urllc}{URLLC}{Ultra-reliable and low latency communication}
\newacronym{evt}{EVT}{Extreme Value Theory}
\newacronym{gpd}{GPD}{Generalized Pareto Distribution} 
\newacronym{cnn}{CNN}{Convolutional Neural Networks} 
\newacronym{gan}{GAN}{Generative Adversarial Networks} 
\newacronym{vae}{VAE}{Variational Autoencoder} 
\newacronym{los}{LoS}{Line-of-Sight} 
\newacronym{nlos}{NLoS}{Non-Line-of-Sight} 
\newacronym{snr}{SNR}{Signal-to-Noise Ratio} 
\newacronym{rmse}{RMSE}{Root Mean Square Error} 
\newacronym{elbo}{ELBO}{Evidence Lower Bound} 
\newacronym{dpm}{DPM}{Dominant Path Model} 
\newacronym{irt}{IRT}{Intelligent Ray Tracing} 
\newacronym{drnn}{DRNN}{Deep Recurrent Neural Network}
\newacronym{gen-ai}{GenAI}{Generative Artificial Intelligence}
\newacronym{ai}{AI}{Artificial Intelligence}
 \newacronym{cdf}{CDF}{Cumulative Distribution Function}
 \newacronym{gmm}{GMM}{Gaussian Mixture Model}

\usepackage{multirow}
\usepackage{algorithm}
\usepackage{algpseudocode}
\usepackage{bbm}
\usepackage{epstopdf}
\usepackage{amsmath,bm}
\usepackage{verbatim}
\usepackage{tikz}
\usetikzlibrary{calc, positioning, shadows, arrows.meta}

\begin{document}

\title{Physics-informed VAE-EVT for Tail Aware Radio Map Prediction}
\thispagestyle{empty}
\author{\IEEEauthorblockN{Amanda Sheron Gamage, Niloofar Mehrnia, and James Gross
}\\
\IEEEauthorblockA{Department of Electrical Engineering and Computer Science, KTH Royal Institute of Technology, Sweden}\\
e-mails:\{asgamage,nilome,jamesgr\}@kth.se
        
\thanks{Niloofar Mehrnia acknowledges the support of  Vinnova (Sweden’s Innovation Agency) under Grant \#2025-01333.}
}

\maketitle
\thispagestyle{empty}

\begin{abstract}
Ultra-reliable low-latency communication (URLLC) requires precise identification of spatial regions where the signal-to-noise ratio (SNR) falls below an outage threshold. In this context, an outage refers to instances in which SNR falls below a specified threshold, which, for URLLC, can be as stringent as the \(0.1\%\) quantile of the SNR distribution. Traditional generative radio map models tend to focus on reconstructing average signal levels, often overlooking the low-SNR that is crucial for accurate outage prediction. To address this limitation, we introduce a physics- and tail-informed VAE–EVT (variational autoencoder–extreme value theory) framework that distinctly models both the bulk and tail distribution of SNR. Our approach begins with a physics-informed preprocessing stage that extracts deterministic features, including line-of-sight, shadowing, and distance, from the scene geometry. A dual-latent encoder then captures the bulk SNR using a Gaussian mixture and the tail using a generalized Pareto distribution (GPD). By employing a modified variational objective, the model is trained to jointly supervise both regimes, ensuring focused attention on extreme fading events. Evaluated on the RadioMapSeer dataset, our method achieves an SNR RMSE of \(4.83\)~dB in the outage region defined by the low threshold of \(0.1\%\) SNR-quantile. This significantly outperforms the state-of-the-art GAN-based model, which records an SNR RMSE of \(21.90\)~dB, with the performance gap widening as the outage threshold becomes more stringent.
\end{abstract}


\section{Introduction}
\label{sec:introduction}

The rapid expansion of mission-critical applications, such as autonomous driving and industrial automation, has placed unprecedented demands on wireless networks, establishing \gls{urllc} as an essential design paradigm for next-generation systems \cite{bennis2018ultrareliable}. \gls{urllc} fundamentally requires outage probabilities as low as $10^{-9}$ to $10^{-5}$. Achieving such stringent reliability at the network level depends on precise radio map predictions, which provide detailed spatial insights into signal quality and serve as the geographic basis for guaranteeing reliability \cite{zeng2024tutorial}.  Identifying the spatial locations where signal strength drops below a threshold is vital, as these infrequent but severe fades largely determine overall reliability. Traditional model-based methods often fall short in capturing the intricate, site-specific fading characteristics anticipated in 6G environments. Consequently, data-driven approaches, particularly those leveraging \gls{gen-ai} models, have become indispensable for accurately modeling outage behavior.

A radio map provides a spatially resolved representation of signal quality across a geographic area, serving as a foundational tool for network planning and proactive resource allocation \cite{zeng2024tutorial}. Although deterministic methods like ray tracing can deliver highly accurate results, they depend on exhaustive, site-specific environmental data, which significantly limits their scalability in varied and rapidly changing settings. This limitation has driven sustained interest in learning-based alternatives, such as RadioUNet \cite{levie2021radiounet}, which demonstrated that generative-based encoder-decoder networks can efficiently reconstruct path loss maps from building geometry. However, a critical gap remains where most advanced models focus primarily on reconstructing average statistics and often overlook spatially deep fades in the signal tail. For \gls{urllc} applications, relying solely on average signal statistics is inadequate; instead, it is essential to explicitly model the rare but critical extremes that ultimately determine system reliability.

\gls{evt} provides a statistical foundation for characterizing rare fading events via the \gls{gpd}. The authors in \cite{Mehrnia_2021, Mehrnia_2022} have conducted foundational research on estimating the channel tail distribution for \gls{urllc} by fitting the \gls{gpd} to the extreme quantiles of received signal power. This work is subsequently generalized in \cite{Mehrnia2025ChannelPrediction} to address channel prediction, where an \gls{evt}-based adaptive quantile loss function is introduced within a \gls{drnn} framework. Additionally, the authors in \cite{2024evtenrichedradiomapsurllc} used \gls{evt} for constructing reliability-aware radio maps through Gaussian process interpolation of the \gls{gpd} tail parameters at unobserved locations. However, these approaches lack a generative core for map synthesis and remain tied to the specific environments in which they were trained, requiring retraining when deployed in new settings, and underscoring the need for a more flexible, generative approach to \gls{evt} based radio map synthesis.

\gls{gen-ai} has recently gained attention for large-scale radio map generation. For instance, \cite{zhang_rme_gan_2023} introduced a framework that uses a \gls{gan} to estimate radio maps in complex urban environments, aiming to enhance the quality of the generated maps. \gls{gan}-based approaches are particularly efficient during inference but typically concentrate on modeling the central tendencies of the signal distribution, often neglecting the spatially localized deep fades that are crucial for \gls{urllc}. More recently, diffusion models have shown promise in capturing near-extreme quantiles of the signal distribution, as highlighted in \cite{radiodiff_wang_tao_2025}. Despite their effectiveness, diffusion models are computationally intensive and incur high inference latency. \cite{valiahdi2026evtbasedgenerativeaitailaware} integrates \gls{evt} with \gls{gen-ai} for channel estimation, employing a \gls{gan} to selectively enrich the scarce extreme region and to estimate the tail threshold and \gls{gpd} parameters online. That framework, however, operates on a scalar received power time series at a single link, and therefore characterizes the temporal tail of one channel rather than the spatial distribution of outage across an environment. Extending tail-aware generation from a per-link time series to a spatially resolved radio map additionally requires conditioning on the scene geometry that determines where deep fades occur. Limited fidelity in the low-SNR tail and the high computational cost of tail-aware generation motivate incorporating domain structure beyond purely data-driven generative modeling. 

Physics-informed AI has recently emerged as a powerful approach in wireless communications to address the limitations of purely generative methods. By embedding propagation models, site-specific geometry, or analytical path loss priors into the learning framework, these methods infuse domain knowledge directly into the model training process \cite{chen2024diffractionscatteringawareradio, jaensch2025radiomappredictionaerial}. This physical conditioning enables the network to accurately resolve the transitions in signal power that define outage boundaries, thereby delivering a reliable and interpretable radio map prediction. 



In this paper, we propose a novel Physics-informed VAE-EVT framework that integrates \gls{gen-ai}, \gls{evt}, and physics-informed AI to predict radio maps. The generative model learns the bulk and extreme quantiles (tails) of the signal distribution separately, while deterministic scene-geometry conditions constrain the prediction to ensure physical consistency.
The main contributions of the paper are as follows:
\begin{itemize}
    \item We develop a physics-informed feature extraction pipeline that computes spatial priors, including \gls{los} and \gls{nlos} segregation, shadow depth, and distance attenuation, to directly encode the physics of propagation into the model's input.
    \item We introduce a novel \emph{Dual Latent} Encoder architecture that integrates a standard \gls{vae} with \gls{evt}. This explicitly separates the latent space, modeling bulk-signal conditions with a standard Gaussian distribution and extreme-signal outages with a \gls{gpd}.
    \item We evaluate the performance of our framework compared to the state-of-the-art algorithms, using the RadioMapSeer dataset \cite{DatasetPaper}, a publicly available benchmark widely used for radio map generation.
\end{itemize}

The remainder of this paper is structured as follows. Section~\ref{sec:system_model} outlines the system model assumptions. In Section~\ref{sec:method}, we detail the proposed Physics-informed VAE-EVT framework. Section~\ref{sec:numerical_results} discusses the measurement data and model assumptions, and provides a comprehensive evaluation. Finally, Section~\ref{sec:conclusions} concludes the paper.\footnote{The implementation is available at \url{https://github.com/AmandaGamage/physics-informed-vae-evt}.}
\section{System Model and Problem Formulation}
\label{sec:system_model}

We consider an urban radio propagation environment discretized into an $M \times M$ spatial grid. This is represented by a binary building occupancy map $\mathcal{B}(p) \in \{0,1\}^{M \times M}$, where $\mathcal{B}(p)=1$ indicates a building at pixel $p=(x,y)$, and $\mathcal{B}(p)=0$ denotes free space. Given a single transmitter placed at a known location $p_{tx}$, the set of valid receiver locations is defined as $\Omega = \{p: \mathcal{B}(p)=0\}$. Our primary objective is to characterize the spatial distribution of the received \gls{snr} across $\Omega$. We place particular emphasis on accurately predicting the extremely low \gls{snr}s that trigger outage, as capturing these rare events is critical for evaluating the reliability of \gls{urllc} systems.

\subsection{Outage Definition}
\label{sec:signal_model}

Let $P_{rx}(p)$ denote the received power in dBm at pixel $p$. For a system with bandwidth $B$, noise spectral density $N_0$, and receiver noise figure $\mathrm{NF}$, the thermal noise floor is calculated as $P_{noise} = 10\log_{10}(B) + N_0 + \mathrm{NF}$~dBm. Then, the \gls{snr} in dB is therefore given by:
\begin{equation}
    \gamma(p) = P_{rx}(p) - P_{noise}.
\end{equation}

An outage is defined as the event where $\{\gamma(p) < \gamma_{th}\}$, for a predefined critical \gls{snr} threshold, $\gamma_{th}$. Consequently, the outage region for a given environment is defined as:
\begin{equation}
    \mathcal{O} = \{p \in \Omega : \gamma(p) < \gamma_{th}\}.
\end{equation}

\subsection{Deterministic Geometric Structure and Residual Variability}
\label{sec:decomposition}

The \gls{snr}, $\gamma(p)$, comprises two distinct components: The \emph{large-scale} component is deterministic and entirely governed by the macroscopic scene geometry, and the \emph{small-scale} component that captures the probabilistic variability arising from complex physical phenomena, such as diffraction, scattering, and signal interactions near building edges. 

The deterministic component can be analytically derived from physical propagation principles. Once $\mathcal{B}$ and $p_{tx}$ are defined, essential geometric metrics, including distance from the transmitter, \gls{los}/\gls{nlos} condition, and localized shadowing, can be explicitly computed via ray tracing \cite{chen2024diffractionscatteringawareradio}. These metrics are assembled into a comprehensive set of deterministic descriptors, denoted as $\mathbf{X}_{geo}$. Then, small-scale statistical variability is modeled as the conditional distribution $p(\gamma \mid \mathbf{X}_{geo})$. 

\subsection{Bulk and Tail of the Conditional SNR Distribution}
\label{sec:bulk_tail_model}

Conditioned on $\mathbf{X}_{geo}$, the \gls{snr} exhibits two inherently distinct statistical regimes. Under standard average-statistics channel conditions (the \emph{bulk}), $\gamma(p)$ is captured by a \gls{gmm}~\cite{selim2016mog}:
\begin{equation}
    f_{bulk}(\bm{\gamma} \, : \gamma(p)>\gamma_{th} \mid \mathbf{X}_{geo} ) =
    \sum_{k=1}^{K} \alpha_k\,
    \mathcal{N}\!\left(\bm{\gamma};\,\mu_k,\,\sigma_k^2\right),
\end{equation}
where $K$ represents the total number of mixture components, $\mathcal{N}$ denotes the Gaussian probability density function, and $\alpha_k$, $\mu_k$, and $\sigma_k^2$ denote the mixing weight, mean, and variance of the $k$-th component, respectively. The mixing weights are strictly constrained such that $\sum_{k=1}^K \alpha_k = 1$. Each mixture component $k$ corresponds to a specific propagation state, such as \gls{los}, \gls{nlos}, or partial obstruction. 

A finite Gaussian mixture systematically underestimates the probability of severe, rare fading events. To overcome this limitation, under tail-statistics channel conditions ($\gamma(p) < \gamma_{th}$), $\gamma(p)$ is modeled using a \gls{gpd} \cite{Mehrnia_2021}. According to the Pickands-Balkema-de Haan theorem, the distribution of exceedances below a sufficiently low threshold $\gamma_{th}$ \cite{coles2001introduction} converges to:
\begin{equation}
    f_{tail}(\bm{\gamma} \, : \gamma(p)<\gamma_{th} \mid \mathbf{X}_{geo} ) = 1 - \left(1 + \dfrac{\xi (\gamma_{th}-{\gamma_{(p)}})}{\beta}\right)^{-1/\xi},
\end{equation}
where $\beta > 0$ and $\xi \in \mathbb{R}$ represent the \emph{scale} and \emph{shape} parameters of the \gls{gpd} fitted to the exceedances $\bm{\gamma} = \gamma_{th}-\gamma(p)$, respectively.

To robustly learn the full conditional distribution $p(\bm{\gamma} \mid \mathbf{X}_{geo})$, we utilize a \gls{vae}~\cite{Kingma2014AutoEncoding} driven by a specialized training objective. The reconstruction loss is two-fold, ensuring the bulk is supervised under a Gaussian assumption while the tail is simultaneously supervised under a \gls{gpd} assumption.

\section{Proposed VAE-EVT Framework for Tail-Aware Radio Map Prediction}
\label{sec:method}

\begin{figure*}[htbp]
    \centering
    \Large
    \resizebox{\textwidth}{!}{
    
    \definecolor{cInputFill}{HTML}{F8F9FA}   
    \definecolor{cInputBorder}{HTML}{6C757D} 
    
    \definecolor{cEncDecFill}{HTML}{E3F2FD}  
    \definecolor{cEncDecBorder}{HTML}{1565C0}
    
    \definecolor{cBulkFill}{HTML}{E0F2F1}    
    \definecolor{cBulkBorder}{HTML}{00695C}  
    
    \definecolor{cTailFill}{HTML}{E8EAF6}    
    \definecolor{cTailBorder}{HTML}{283593}  
    
    \definecolor{cConcatFill}{HTML}{F1F3F5}  
    \definecolor{cConcatBorder}{HTML}{495057}
    
    \definecolor{cPiFill}{HTML}{FFF8E1}      
    \definecolor{cPiBorder}{HTML}{FF8F00}    
    
    \definecolor{cOutFill}{HTML}{FFF3E0}     
    \definecolor{cOutBorder}{HTML}{EF6C00}   

    \definecolor{cArrow}{HTML}{455A64}       
    \definecolor{cRule}{HTML}{CFD8DC}        

    \begin{tikzpicture}[
        font=\Large,
        >=Stealth,
        node distance=1.2cm and 1.6cm,
        base/.style={
            draw, thick, align=center, rounded corners=6pt,
            inner sep=4pt,              
            minimum width=4cm,
            minimum height=2.2cm,       
            drop shadow={opacity=0.15, shadow xshift=0.04cm, shadow yshift=-0.04cm}
        },
        input/.style={base, fill=cInputFill, draw=cInputBorder, dashed, thick},
        encoder/.style={base, fill=cEncDecFill, draw=cEncDecBorder},
        decoder/.style={base, fill=cEncDecFill, draw=cEncDecBorder},
        bulk/.style={base, fill=cBulkFill, draw=cBulkBorder},
        tail/.style={base, fill=cTailFill, draw=cTailBorder},
        pibranch/.style={base, fill=cPiFill, draw=cPiBorder},
        concat/.style={base, fill=cConcatFill, draw=cConcatBorder},
        output/.style={base, fill=cOutFill, draw=cOutBorder},
        arrow/.style={->, thick, draw=cArrow, rounded corners=6pt}
    ]

    \newcommand{\softrule}{\vspace*{-1ex}\color{cRule}\rule{3.6cm}{1.2pt}\color{black}\vspace*{0.5ex}}

    \node[input] (input) {
        \textbf{Input Tensor} \\
        \softrule \\
        $\mathbf{X}_{geo}$
    };

    \node[encoder, right=2cm of input] (enc) {
        \textbf{Dual Latent} \\
        \textbf{Encoder}
    };

    \node[bulk, right=1.6cm of enc, yshift=2cm] (z_bulk) {
        \textbf{Bulk Latent} \\
        \softrule \\
        $\mathbf{z}_{bulk} \sim \mathcal{N}(\mu_g, \sigma_g^2 \mathbf{I})$
    };

    \node[tail, right=1.6cm of enc, yshift=-2cm] (z_tail) {
        \textbf{Tail Latent} \\
        \softrule \\
        $\mathbf{z}_{tail} \sim \text{GPD}(\hat{\xi}, \hat{\beta})$       
    };

    \node[concat] (concat) at ($(z_bulk)!0.5!(z_tail) + (6.8cm, 0)$) {
        \textbf{Concatenation} \\
        \softrule \\
        $[\mathbf{z}_{bulk}; \mathbf{z}_{tail}] \rightarrow \text{FC}$
    };

    \node[decoder, right=2cm of concat] (dec) {
        \textbf{U-Net Decoder} \\
        \softrule \\
        Spatial Attention
    };

    \node[bulk, right=1.6cm of dec, yshift=2.7cm] (out_bulk) {
        \textbf{Bulk Branch} \\
        \softrule \\
        $\mu$, $\log\sigma^2$
    };

    \node[pibranch, right=1.6cm of dec, yshift=0cm] (out_pi) {
        \textbf{Outage Branch} \\
        \softrule \\
        $\pi \rightarrow \pi_s$
    };

    \node[tail, right=1.6cm of dec, yshift=-2.7cm] (out_tail) {
        \textbf{Tail Branch} \\
        \softrule \\
        $y_t$
    };

    \node[output, right=2.8cm of out_pi] (final) {
        \textbf{Final Output} \\
        \vspace*{-1ex}\color{cRule}\rule{4.5cm}{1.2pt}\color{black}\vspace*{0.5ex} \\
        $\hat{\gamma} = (1 - \pi_s)\mu + \pi_s y_t$
    };

    \draw[arrow] (input.east) -- (enc.west);

    \draw[arrow] (enc.east) -- ++(0.7,0) |- (z_bulk.west);
    \draw[arrow] (enc.east) -- ++(0.7,0) |- (z_tail.west);

    \draw[arrow] (z_bulk.east) -- ++(0.5,0) |- ([yshift=0.2cm]concat.west);
    \draw[arrow] (z_tail.east) -- ++(0.7,0) |- ([yshift=-0.2cm]concat.west);

    \draw[arrow] (concat.east) -- (dec.west);

    \draw[arrow] (dec.east) -- ++(0.7,0) |- (out_bulk.west);
    \draw[arrow] (dec.east) -- (out_pi.west);
    \draw[arrow] (dec.east) -- ++(0.7,0) |- (out_tail.west);

    \draw[arrow] (out_bulk.east) -- ++(0.7,0) |- ([yshift=0.4cm]final.west);
    \draw[arrow] (out_pi.east) -- (final.west);
    \draw[arrow] (out_tail.east) -- ++(0.7,0) |- ([yshift=-0.4cm]final.west);

    \end{tikzpicture}
    }
    \caption{Overview of the proposed Physics-informed VAE-EVT architecture.}
    \label{fig:model_architecture}
\end{figure*}
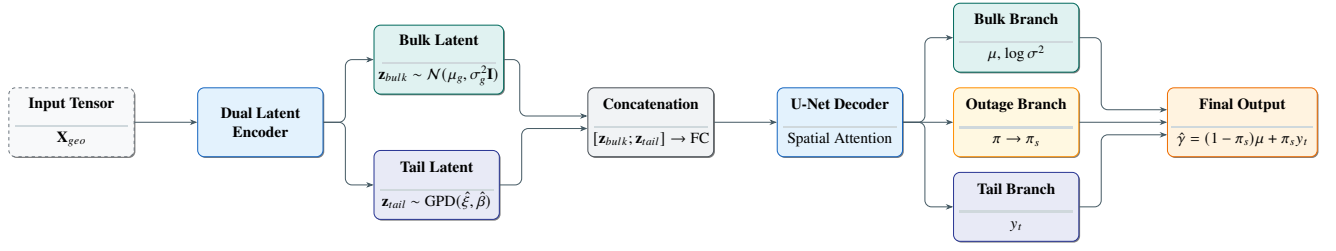

The proposed Physics-informed VAE-EVT framework (shown in Fig.~\ref{fig:model_architecture}) comprises two core elements: (i) a physics-informed preprocessing pipeline that extracts the deterministic structural tensor $\mathbf{X}_{geo}$, and (ii) a modified \gls{vae} architecture featuring a dual-latent encoder and a tail-aware training objective. Together, these elements enable the simultaneous learning of the bulk and tail regimes to accurately estimate $p(\bm{\gamma}\mid \mathbf{X}_{geo})$. Physics-informed preprocessing produces the multi-channel input tensor \(\mathbf{X}_{geo}\) (\textit{Input Tensor}). A \textit{Dual Latent Encoder} maps \(\mathbf{X}_{geo}\) to two latent variables: a Gaussian bulk latent \(\mathbf{z}_{bulk}\) and a GPD-anchored tail latent \(\mathbf{z}_{tail}\) (\textit{Bulk Latent}, \textit{Tail Latent}). The two latents are concatenated and passed through a fully connected layer (\textit{Concatenation}). A \textit{U-Net Decoder} with spatial attention then reconstructs spatial feature maps and feeds three output branches: a \textit{Bulk Branch} producing \(\mu\) and \(\log\sigma^2\), a \textit{Tail Branch} producing \(y_t\), and an \textit{Outage Branch} producing \(\pi\) that is sharpened to \(\pi_s\), all used to estimate SNR, $\hat{\gamma}$, as the \textit{Final Output}.

\subsection{Physics-informed Preprocessing}
\label{sec:preprocessing}

The preprocessing stage converts \((\mathcal{B},p_{tx})\) into a tensor \(\mathbf{X}_{geo}\) of deterministic geometric descriptors. All channels are defined on the full \(M\times M\) grid; for building pixels (\(\mathcal{B}(p)=1\)), channels are set to 0 and masked during training and evaluation.

\subsubsection{Distance and Transmitter Proximity}

For each pixel \(p\in\Omega\), the Euclidean distance to the transmitter is \(d(p)=\|p-p_{tx}\|_2\).
We construct a min--max normalized log-distance channel \(D_{norm}(p)\) = \(\log(d(p)+\epsilon_d)\) with a small \(\epsilon_d>0\) to avoid \(\log 0\). We also construct a linearly normalized distance channel \(D_{all}(p)\) over \(\Omega\), used as a stable conditioning input for the tail and outage branches.

A smooth proximity map \(T_x(p)\) is defined as a truncated Gaussian centered at \(p_{tx}\), \(T_x(p)=\exp\!(-\frac{\|p-p_{tx}\|_2^2}{2\sigma_T^2})\), with \(\sigma_T>0\) chosen relative to the grid scale.

\subsubsection{\gls{los} Mask and \gls{nlos} Penetration Depth}

The binary \gls{los} mask \(M_{LOS}(p)\in\{0,1\}\) is set to 1 when the discretized ray segment from \(p_{tx}\) to \(p\) does not intersect any building pixel in \(\mathcal{B}\), and 0 otherwise. Let \(n_{hits}(p)\in\mathbb{N}\) be the number of building intersections along this ray under the chosen discretization rule.
The LoS-masked log-distance feature is
\begin{equation}
d_L(p)=D_{norm}(p)\,M_{LOS}(p).
\end{equation}
For NLoS pixels, the obstruction severity is captured by the normalized penetration depth
\begin{equation}
D_{NLOS}(p)=\bigl(1-M_{LOS}(p)\bigr)\cdot \min\!\left(\frac{n_{hits}(p)}{d_{max}},\,1\right),
\end{equation}
where \(d_{max}\in\mathbb{N}\) is a fixed normalization constant (e.g., the maximum \(n_{hits}\) observed in the training set).

\subsubsection{Localized Shadowing and Edge Map}

A binary \gls{los} mask does not capture partial shadowing near corners and corridors. We define a localized shadowing score \(S_{NLOS}(p)\in[0,1]\) by sampling \(N_s\) equally spaced points along the segment from \(p\) toward \(p_{tx}\), and computing the fraction of samples that fall inside buildings. We set \(S_{NLOS}(p)=0\) for LoS pixels.

To provide a guidance signal for spatial attention, we compute an edge map \(E(p)\) emphasizing LoS/NLoS boundaries and strong shadow transitions. Let \(\nabla\) denote the image gradient estimated with Sobel operators, and let \(G_{\sigma_E}\) denote a Gaussian smoothing kernel with standard deviation \(\sigma_E\). We form \(E(p)=\mathrm{clip}\,(
\mathrm{norm}\!(
\alpha_1 \|\nabla M_{LOS}\|(p) + \alpha_2 \left\|\nabla\!\bigl(G_{\sigma_E} * S_{NLOS}\bigr)\right\|(p)
))\), where \(\alpha_1,\alpha_2\ge 0\) are weighting coefficients, \(\mathrm{norm}(\cdot)\) denotes a fixed normalization (e.g., min--max over \(\Omega\)), and \(\mathrm{clip}(\cdot)\) truncates the result to \([0,1]\).

\subsubsection{Outage Prior}

A coarse geometric baseline for outage risk is provided through an outage prior:
\begin{equation}
\label{eq:outage_prior}
P_{outage}(p)=
\begin{cases}
\mathrm{clip}_{(\epsilon,\,1-\epsilon)}\!\bigl(a_1 D_{norm}(p)\bigr), & M_{LOS}(p)=1, \\[4pt]
\mathrm{clip}_{(\epsilon,\,1-\epsilon)}\!\bigl(b_1 S_{NLOS}(p)+b_2 D_{NLOS}(p)+b_3 D_{norm}(p)\bigr), & M_{LOS}(p)=0,
\end{cases}
\end{equation}
where \(a_1,b_1,b_2,b_3\ge 0\) are coefficients and \(\epsilon\in(0,0.5)\) avoids degenerate probabilities.

\subsubsection{Per-map Threshold and Input Tensor}

For each environment (radio map), \(\gamma_{th}\) is set as the \(q\)-th percentile of the SNR values \(\{\gamma(p):p\in\Omega\}\), where \(q\in(0,100)\) is fixed across the dataset. A normalized threshold encoding \(\hat{\gamma}_{th}\) is then broadcast to all pixels as an additional channel. Concretely, \(\hat{\gamma}_{th}\) can be formed using dataset-level normalization constants \(\gamma_{\min}\) and \(\gamma_{\max}\) (computed on the training set):
\(\hat{\gamma}_{th}=\frac{\gamma_{th}-\gamma_{\min}}{\gamma_{\max}-\gamma_{\min}+\epsilon_\gamma}\), with \(\epsilon_\gamma>0\) preventing division by zero.

To anchor the tail latent space, dataset-level GPD parameters \((\hat{\xi},\hat{\beta})\) are estimated by maximum likelihood using training-set exceedances, \(\gamma_{th}-\gamma\), restricted to pixels with \(\gamma<\gamma_{th}\). These anchors are treated as constants during network training.
The final input tensor is \(\mathbf{X}_{geo}=\mathrm{stack}(
\mathcal{B},\, T_x,\, M_{LOS},\, d_L,\, S_{NLOS},\, D_{NLOS},\, E,\, P_{outage},\, D_{all}, \newline \hat{\gamma}_{th})\), and corresponds to the \textit{Input Tensor} block in Fig.~\ref{fig:model_architecture}.

\subsection{Dual-Latent Encoder}
\label{sec:encoder}

The encoder implements the approximate posterior \(q_\phi(\mathbf{z}_{bulk},\mathbf{z}_{tail}\mid \mathbf{X}_{geo})\) shown as the \textit{Dual Latent Encoder} block in Fig.~\ref{fig:model_architecture}. It uses a convolutional backbone to produce multi-resolution feature maps, which are cached for the U-Net skip connections in the decoder.

At the bottleneck, two dense heads produce parameters for the bulk latent and for an auxiliary tail pathway that is mapped to a GPD-anchored latent.

\subsubsection{Bulk Latent Variable, \(\mathbf{z}_{bulk}\)}

The bulk latent is Gaussian with diagonal covariance. The encoder outputs \(\mu_g\in\mathbb{R}^{d_z}\) and \(\log\sigma_g^2\in\mathbb{R}^{d_z}\), where \(d_z\) is the bulk latent dimension, and sampling is performed via reparameterization. \(\mathbf{z}_{bulk}=\mu_g+\sigma_g\odot \boldsymbol{\epsilon}\), where \(\boldsymbol{\epsilon}\sim\mathcal{N}(\mathbf{0},\mathbf{I})\), \(\sigma_g=\exp\!\left(\tfrac{1}{2}\log\sigma_g^2\right)\), and \(\odot\) denoting elementwise multiplication. This matches the \textit{Bulk Latent} block \(\mathbf{z}_{bulk}\sim\mathcal{N}(\mu_g,\sigma_g^2\mathbf{I})\) in the diagram.

\subsubsection{Tail Latent Variable, \(\mathbf{z}_{tail}\)}

The tail pathway begins with an auxiliary diagonal-Gaussian variable. The encoder outputs \(\mu_p\in\mathbb{R}^{d_t}\) and \(\log\sigma_p^2\in\mathbb{R}^{d_t}\), where \(d_t\) is the tail latent dimension, and draws \(\mathbf{z}_{aux}=\mu_p+\sigma_p\odot \boldsymbol{\epsilon}\), where, \(\boldsymbol{\epsilon}\sim\mathcal{N}(\mathbf{0},\mathbf{I})\), and \(\sigma_p=\exp\!(\tfrac{1}{2}\log\sigma_p^2)\).
A differentiable approximation of a uniform random variable is then obtained by \(\mathbf{u}=\sigma(0.5*\mathbf{z}_{aux}) \in (0,1)^{d_t}\), where \(\sigma(\cdot)\) is the logistic sigmoid.

Finally, \(\mathbf{u}\) is mapped through the GPD quantile function using the fixed anchors \((\hat{\xi},\hat{\beta})\). Accordingly, \(\mathbf{z}_{tail}
=\frac{\hat{\beta}}{\hat{\xi}}((1-\mathbf{u})^{-\hat{\xi}}-\mathbf{1})\), with the exponent applied elementwise. In practice, when \(|\hat{\xi}|\) is very small, the numerically stable exponential-limit form can be used:
\(\mathbf{z}_{tail}\approx -\hat{\beta}\log(1-\mathbf{u})\).

\subsection{Concatenation and U-Net Decoder with Spatial Attention}
\label{sec:decoder}

The two latents are concatenated exactly as shown in the \textit{Concatenation} block, \(\mathbf{z}=[\mathbf{z}_{bulk};\,\mathbf{z}_{tail}]\), then passed through a fully connected (FC) layer and reshaped into a low-resolution feature map that seeds the decoder. The decoder follows a U-Net design: features are progressively upsampled back to \(M\times M\), concatenated with encoder features at matching resolutions through skip connections, and refined with convolutional residual blocks. A spatial attention module is applied within the decoder to emphasize spatially localized transitions highlighted by channels such as \(E\) and \(S_{NLOS}\).

\subsection{Output Heads and Final Prediction}
\label{sec:heads}

From the shared decoder feature map, three heads produce the outputs shown on the right of Fig.~\ref{fig:model_architecture}. Let \(m(p)=1-\mathcal{B}(p)\) be the free-space mask, used to force building pixels to be excluded from reconstruction losses.

\subsubsection{Bulk Branch}

The bulk branch outputs \(\mu(p)\) and \(\log\sigma^2(p)\) for each pixel. These define a Gaussian likelihood for SNR reconstruction in the non-outage regime.

\subsubsection{Tail Branch}

The tail branch outputs \(y_t(p)\), a tail-focused SNR estimate intended for outage pixels. To keep the tail head aligned with the shortfall interpretation, the model may internally parameterize a nonnegative shortfall \(\hat{y}(p)\ge 0\) and set \(y_t(p)=\gamma_{th}-\hat{y}(p)\), so that larger shortfalls correspond to smaller tail SNR values.

\subsubsection{Outage Branch}

The outage branch predicts a per-pixel outage probability \(\pi(p)\in[0,1]\). Building pixels are masked by setting \(\pi(p)\leftarrow m(p)\,\pi(p)\).
A sharpened routing mask \(\pi_s(p)\) is obtained using a steep sigmoid, \(\pi_s(p)=\sigma\!\big(\kappa(\pi(p)-t^*)\big)\), where \(\kappa>0\) controls the steepness and \(t^*\in(0,1)\) is selected on a validation set by maximizing the F1 score. This corresponds to \(\pi\rightarrow\pi_s\) operation in Fig.~\ref{fig:model_architecture}.

\subsubsection{Final Output}

The final SNR prediction is computed per pixel as \(\hat{\gamma}(p)=(1-\pi_s(p))\,\mu(p)+\pi_s(p)\,y_t(p)\), matching the \textit{Final Output} block in Fig.~\ref{fig:model_architecture}.

\subsection{VAE-EVT Training Loss}
\label{sec:loss}

The model is trained end-to-end using a combination of reconstruction, latent regularization, and explicit supervision for outage routing. The total objective is
\begin{equation}
\begin{split}
\mathcal{L}_{total}
&= \lambda_r\mathcal{L}_{recon}
+ \lambda_{KL,b}\mathcal{L}_{KL,b}
+ \lambda_{KL,t}\mathcal{L}_{KL,t}\\
&\quad + \lambda_\pi\mathcal{L}_\pi
+ \lambda_{out}\mathcal{L}_{outage}
 + \lambda_s \mathcal{L}_{sharp}+ \lambda_{LoS}\mathcal{L}_{LoS},
\end{split}
\end{equation}
where \(\lambda_{\cdot}\ge 0\) are scalar weights and fine-tuned during training.

\subsubsection{Reconstruction Loss \(\mathcal{L}_{recon}\)}

Reconstruction is routed by the predicted outage probability so that bulk learning dominates when \(\pi\) is small and tail learning dominates when \(\pi\) is large,
\(\mathcal{L}_{recon}
=\frac{1}{|\Omega|}\sum_{p\in\Omega}
\left[(1-\pi(p))\,\mathcal{L}_{NLL}(p)+\pi(p)\,\bigl(y_t(p)-\gamma(p)\bigr)^2\right]\).
The bulk negative log-likelihood term is \(\mathcal{L}_{NLL}(p)
=\tfrac{1}{2}\left(\log\!\big(2\pi\sigma^2(p)\big)+\frac{(\gamma(p)-\mu(p))^2}{\sigma^2(p)}\right)\), where, \(\sigma^2(p)=\exp\!\big(\log\sigma^2(p)\big)\).

\subsubsection{Outage-Region Emphasis \(\mathcal{L}_{outage}\)}

Because outage pixels are rare, \(\mathcal{L}_{outage}\) increases the penalty inside
\(\mathcal{O}=\{p\in\Omega:\gamma(p)\le\gamma_{th}\}\). Accordingly, \(\mathcal{L}_{outage}
=\frac{1}{|\mathcal{O}|+\epsilon_O}\sum_{p\in\Omega}\mathbb{I}\{p\in\mathcal{O}\}\,\bigl(\hat{\gamma}(p)-\gamma(p)\bigr)^2\), where \(\epsilon_O>0\) prevents division by zero when \(\mathcal{O}\) is empty.

\subsubsection{Latent Regularization \(\mathcal{L}_{KL,b}\) and \(\mathcal{L}_{KL,t}\)}

The bulk KL term regularizes the bulk posterior toward a standard normal prior \(\mathcal{L}_{KL,b}=\mathrm{KL}\!\left(\mathcal{N}(\mu_g,\sigma_g^2\mathbf{I})\,\|\,\mathcal{N}(\mathbf{0},\mathbf{I})\right)\).

For the tail pathway, \(\mathbf{z}_{tail}\) is a deterministic transform of \(\mathbf{z}_{aux}\). We therefore regularize the auxiliary Gaussian, \(\mathcal{L}_{KL,t}=\mathrm{KL}\!\left(\mathcal{N}(\mu_p,\sigma_p^2\mathbf{I})\,\|\,\mathcal{N}(\mathbf{0},\mathbf{I})\right)\), which stabilizes tail sampling while preserving the GPD mapping to \(\mathbf{z}_{tail}\).

\subsubsection{Outage Supervision \(\mathcal{L}_{\pi}\)}

The outage loss supervises \(\pi(p)\) using labels
\(\ell(p)=\mathbb{I}\{\gamma(p)<\gamma_{th}\}, \qquad p\in\Omega\), after masking building pixels by \(m(p)\). In practice, \(\mathcal{L}_{\pi}\) can combine a pixelwise classification loss (e.g., focal or weighted cross-entropy) with region-level overlap and calibration terms. All components operate on \(\pi\) (not \(\pi_s\)) to adjust gradients.

\subsubsection{Sharpening Regularization \(\mathcal{L}_{sharp}\)}

To discourage indecisive routing, \(\mathcal{L}_{sharp}\) penalizes high-entropy outage probabilities, \(\mathcal{L}_{sharp}
=\frac{1}{|\Omega|}\sum_{p\in\Omega}
[-\pi(p)\log(\pi(p)+\epsilon_\pi)-(1-\pi(p))\log(1-\pi(p)+ \epsilon_\pi)]\),
with a small \(\epsilon_\pi>0\) for numerical stability. This term encourages \(\pi\) (and therefore \(\pi_s\)) to approach near-binary routing decisions.

\subsubsection{\gls{los} Outage Emphasis $\mathcal{L}_{LoS}$}
\gls{nlos} outages are well explained by the obstruction descriptors $S_{NLOS}$ and $D_{NLOS}$, whereas a \gls{los} pixel in outage has an unobstructed path and fails nonetheless, through distance attenuation rather than blockage. Since the outage prior of \eqref{eq:outage_prior} and the shadowing channels all key on obstruction, a loss averaged uniformly over $\mathcal{O}$ is dominated by the \gls{nlos} majority, and the tail head converges to a shadowing-driven representation that systematically mispredicts the \gls{los} minority. 
We therefore add an explicit term over the LoS outage subset
$\mathcal{O}_{LoS} = \{p \in \mathcal{O} : M_{LOS}(p) = 1\}$,  
\(\mathcal{L}_{LoS} = \frac{1}{|\mathcal{O}_{LoS}|+\epsilon_{O}}
  \sum_{p \in \Omega} \mathbb{I}\{p \in \mathcal{O}_{LoS}\}
  \Bigl[
    \bigl(\hat{\gamma}(p)-\gamma(p)\bigr)^{2} +
    \bigl(y_{t}(p)-\gamma(p)\bigr)^{2}
\Bigr]\).
This term is the counterpart in the objective to the $D_{all}$ channel in the input tensor. Both exist so that the tail and outage branches can express a distance-driven outage mechanism that the \gls{los} masked geometric channels alone cannot represent.

\section{Numerical Results}
\label{sec:numerical_results}

\subsection{Experimental Setup}
\label{sec:setup}

\subsubsection{Dataset}
\label{sec:dataset}

We evaluate our framework on the RadioMapSeer dataset~\cite{DatasetPaper}, which comprises 700 urban building
layouts drawn from OpenStreetMap~\cite{OSM2017Planet}. Each layout covers a $256 \times 256$~m$^2$ grid at $1$~m resolution, so that $M = 256$, with transmitter, receiver, and building heights of $1.5$~m, $1.5$~m, and $25$~m, respectively. We use the Dominant Path Model maps~\cite{Wahl2005Dominant} simulated with WinProp
, and restricted to $N_{map} = 300$ layouts with $N_{tx} = 10$ transmitter positions each, yielding 3000 environment transmitter pairs split 80/20 into training and test sets. 
The noise floor is computed with $B = 10$~MHz, $N_0 = -174$~dBm/Hz, and $\mathrm{NF} = 0$~dB, giving
$P_{noise} = -104$~dBm. The \gls{snr} values are normalized to $[0,1]$.

\subsubsection{Implementation Details}
\label{sec:impl_details}

The \gls{los} ray tracer uses $100$ samples per ray, sufficient to keep the inter-sample spacing below one pixel along the longest possible diagonal of a $256 \times 256$ map. The localized shadow feature uses $30$ samples,  since it only needs to resolve obstruction in the immediate neighborhood of $p$ rather than the full path to $p_{tx}$. The \gls{nlos} penetration depth is normalized by $d_{max} = 10$, corresponding to the 99th percentile of observed wall crossings in the training set. 
The encoder uses four strided convolutional residual blocks with channel widths $\{32,64,128,256\}$, compressing the input to a $16\times 16$ bottleneck. 
The post-hoc outage sharpening sigmoid uses a slope of $20$, which helps transitions $\pi_s$ from near-zero to near-one within a narrow window. 
Loss weights follow a KL warmup to prevent posterior collapse, with progressive ramping of the classification, outage reconstruction, and sharpening terms during training.

\subsection{Baselines Models}
\label{sec:baselines}
To isolate the contribution of each architectural component, we compare four models, including: (i) Physics-informed VAE-EVT: the proposed model, with dual latent \gls{evt} encoder, physics preprocessing, and explicit \gls{los}/\gls{nlos} feature routing; (ii) VAE-EVT: dual-latent \gls{evt} encoder with no physics preprocessing; (iii) Physics-informed VAE: a standard single-latent \gls{vae} with physics preprocessing but no \gls{evt} tail modeling; and (iv) RME-GAN~\cite{zhang_rme_gan_2023}: a GAN-based radio map estimator with no explicit tail modeling. 

All models are trained using 2400 maps in the training set and are evaluated at two outage thresholds, $q \in \{0.1\%, 10\%\}$. The evaluation metrics include outage region \gls{rmse}, F1-score, Precision, and Recall for outage pixel classification. The outage region \gls{rmse} directly measures tail prediction fidelity, while F1-score captures spatial outage localization accuracy. Results are reported across 600 test maps.

\begin{figure}[!htbp]
    \centering
    \subfloat[]{%
        \includegraphics[width=0.75\columnwidth]{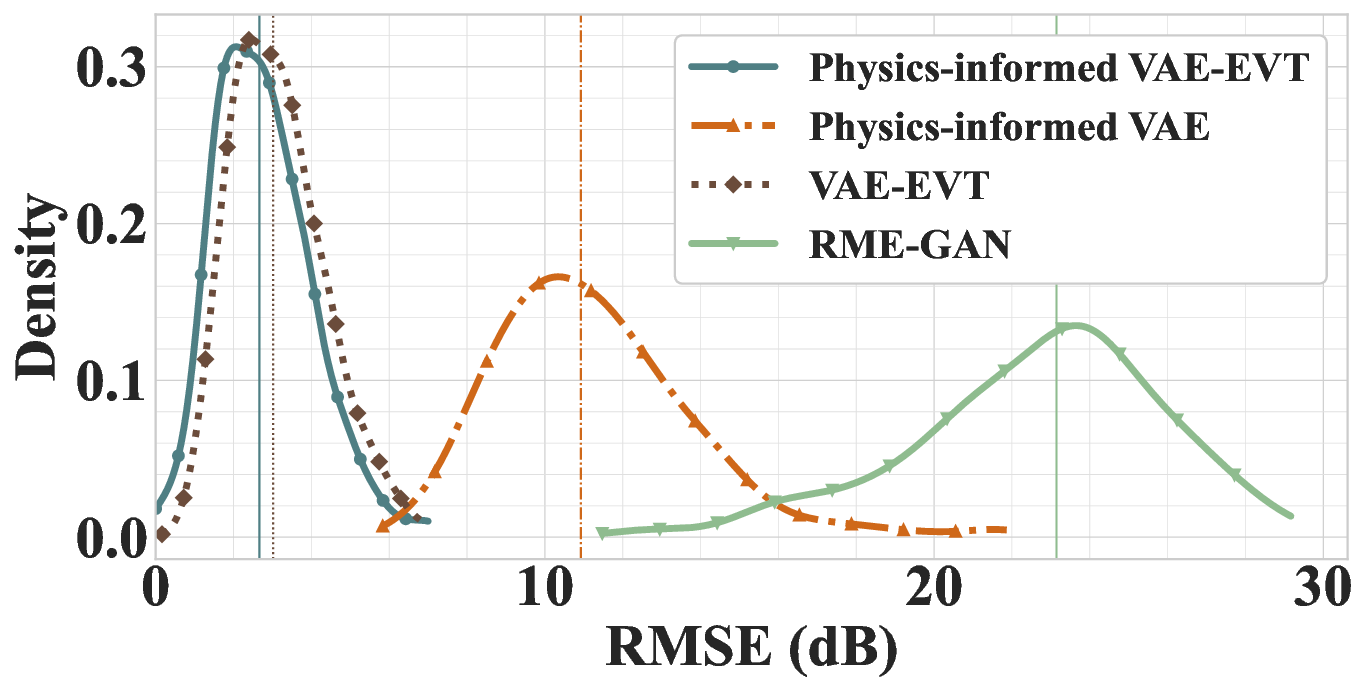}%
        \label{fig:b}
    } \\
    \subfloat[]{%
        \includegraphics[width=0.75\columnwidth]{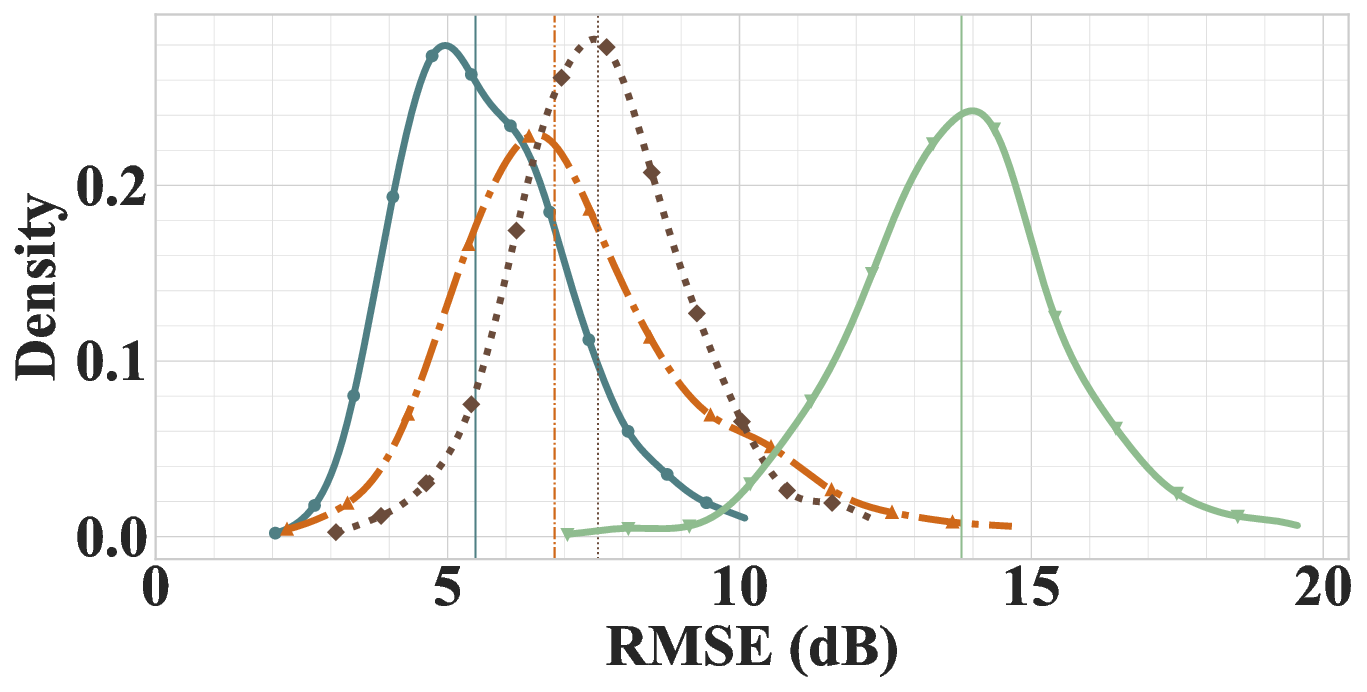}%
        \label{fig:c}
    }
    \caption{Distribution of outage \gls{rmse} across 600 test maps at (a) 0.1\% threshold. (b) 10\% threshold (Legends are common).}
    \label{fig:rmse_all}
\end{figure}

\subsection{Discussion}
\label{sec:main_results}
\begin{figure}[t]
    \centering
    \includegraphics[width=\columnwidth]{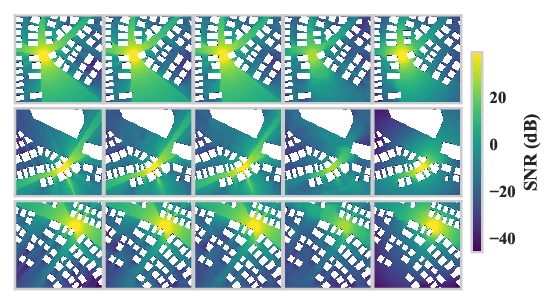}
    \caption{Radio map for three test maps. From left to right: ground truth, Physics-informed VAE-EVT$^*$, Physics-informed VAE, VAE-EVT, and RME-GAN.}
    \label{fig:radio_map_comparison}
\end{figure}

Figure~\ref{fig:radio_map_comparison} shows predicted radio maps for three test maps. RME-GAN produces plausible large-scale propagation patterns but fails to reproduce the fine-grained shadow structure at building boundaries. VAE-EVT misses the high \gls{snr} near the transmitter region entirely. The proposed Physics-informed VAE-EVT most closely matches the ground truth, reproducing both the bright near-transmitter and the deep shadow regions.

\begin{table*}[t]
\centering
\footnotesize
\caption{Comparison of model performance across different thresholds for \emph{outage prediction}.}
\label{tab:model_comparison_combined}
\resizebox{\textwidth}{!}{%
\begin{tabular}{|p{4.5cm}|c|c|c|c|c|c|c|c|c|}
\hline
\multirow{2}{*}{\textbf{Model}} & \multicolumn{2}{c|}{\textbf{RMSE (dB)} } & \multicolumn{2}{c|}{\textbf{F1-Score}} & \multicolumn{2}{c|}{\textbf{Precision} } & \multicolumn{2}{c|}{\textbf{Recall}} \\
\cline{2-9}
 & \textbf{0.1\%}  & \textbf{10\%} & \textbf{0.1\%} & \textbf{10\%} & \textbf{0.1\%}  & \textbf{10\%} & \textbf{0.1\%} & \textbf{10\%} \\
\hline
Physics-informed VAE-EVT$^*$ & 4.83 &  6.97 & 0.1120  & 0.2459 & 0.0608 & 0.8527 & 0.7088 & 0.1437 \\
\hline
Physics-informed VAE & 12.03 & 7.54 & 0.1015 & 0.2903 & 0.0559 & 0.2205 & 0.5504  & 0.6236 \\
\hline
VAE-EVT & 6.56 & 7.81 & 0.1222 & 0.2147 & 0.0673  & 0.9434 & 0.6638  & 0.1211 \\
\hline
RME-GAN \cite{zhang_rme_gan_2023} & 21.90 & 11.53 & 0.1179 & 0.5245 & 0.2524 & 0.5194 & 0.0769 & 0.5297 \\
\hline
\end{tabular}%
}
\end{table*}
Table~\ref{tab:model_comparison_combined} summarizes the performance of different models for two different outage thresholds $q \in \{ 0.1\%,10\%\}$. At the 0.1\% threshold, Physics-informed VAE-EVT achieves 4.83 dB outage \gls{rmse}, outperforming Physics-informed VAE (12.03 dB) by more than 7 dB and RME-GAN (21.90 dB) by roughly 17 dB. The standalone VAE-EVT also remains competitive at 6.56 dB, indicating that only the \gls{evt}-equipped models maintain predictive accuracy at these extreme outage levels. At the 10\% threshold, however, the gap among all models narrows substantially: Physics-informed VAE-EVT records 6.97 dB, while Physics-informed VAE and VAE-EVT reach 7.54 dB and 7.81 dB, respectively, and even RME-GAN improves to 11.53 dB. This compression occurs because the evaluation at 10\% includes a larger number of bulk-similar samples in the tail. Consequently, both the \gls{gpd} fit and the RMSE metric become less indicative of extreme value prediction at this threshold, as they are increasingly influenced by non-extreme values. Although the F1 score at the 0.1\% threshold appears low, the outage \gls{rmse} of $\hat{\gamma}$ remains low because the soft routing through $\pi_s$ combined with $y_t$ keeps misrouted pixels close to the true threshold \gls{snr}, so classification errors translate into only small \gls{rmse} penalties.

Figure~\ref{fig:rmse_all} illustrates the distribution of outage \gls{rmse} across all 600 test maps at the 0.1\% and 10\% thresholds. The distributions corresponding to the \gls{evt}-based models exhibit lower means and variances compared to the non-\gls{evt} models, including Physics-informed VAE or RME-GAN prediction, especially for the 0.1\% outage threshold. This separation confirms that the \gls{evt} advantage is systematic across environments and not driven by a small number of favorable maps. At the 10\% threshold, the distributions overlap, again consistent with the expectation that \gls{evt} provides diminishing returns outside the true extreme tail. 

\begin{figure}[t]
    \centering
    \includegraphics[width=\columnwidth]{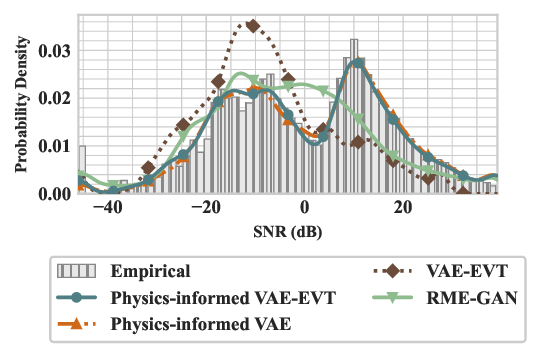}
    \caption{SNR distributions for a sample test map.}
    \label{fig:pdf_comparison}
\end{figure}

Figure~\ref{fig:pdf_comparison} compares the predicted and empirical \gls{snr} densities. The figure reveals a bimodal ground truth, with a dominant \gls{los} peak and a secondary \gls{nlos} peak reflecting the separation between \gls{los}-dominated pixels near the transmitter and deeply shadowed \gls{nlos} pixels behind buildings. In the bulk region, all physics-conditioned \gls{vae} models track the empirical curve closely, while RME-GAN shows a visible offset and VAE-EVT drifts above the empirical curve across the mid-\gls{snr} range. The crucial difference appears in the lower tail (inset). 
The proposed Physics-informed VAE-EVT tracks the empirical densities across both the bulk and the tail, underscoring the importance of physics conditioning. Moreover, models with physics preprocessing reproduce both peaks, while VAE-EVT collapses them and misses the \gls{los} peak entirely.

It is worth mentioning that the original RME-GAN evaluation reported in \cite{zhang_rme_gan_2023} uses all 700 RadioMapSeer regions, split into 500 training, 100 test, and 100 validation regions, whereas every model reported here is trained on the identical subset of $N_{map}=300$ layouts with $N_{tx}=10$ transmitter positions each. We further note that RME-GAN uses 1\% of the true-SNR pixels as sparse observations, whereas the proposed model predicts from scene geometry alone.

The results confirm that \gls{evt} is necessary to capture tail shape, and physics conditioning is necessary to position the distribution correctly along the \gls{snr} and preserve the bimodal bulk structure. It is also worth noting that, in contrast to diffusion-based radio map models that require iterative denoising at inference~\cite{radiodiff_wang_tao_2025}, the proposed framework generates a full $256\times 256$  \gls{snr} map in a single forward pass, making it directly deployable in \gls{urllc} scenarios.
The proposed Physics-informed VAE-EVT, Physics-informed VAE, and VAE-EVT have comparable parameter counts of 38011962, 37882945, and 38000265, respectively. 

\section{Conclusions}
\label{sec:conclusions}

In this paper, we proposed a Physics-informed VAE-EVT framework for tail-aware radio map prediction. By decoupling bulk and tail via a dual-latent encoder and embedding deterministic \gls{los}/\gls{nlos} features, the model achieves 4.83 dB outage \gls{rmse} at the \gls{urllc} critical 0.1\% quantile threshold, compared to 12.03 dB for a Physics-informed VAE and 21.90 dB for the state-of-the-art RME-GAN. The \gls{rmse} distributions confirm that the advantage is systematic across all 600 test environments. These results show that \gls{urllc} grade outage characterization requires both physics-grounded features and principled tail modeling. Future work will extend the framework to multi transmitter scenarios and moving obstruction and receiver targets.

\balance 

\ifCLASSOPTIONcaptionsoff
  \newpage
\fi
\bibliographystyle{ieeetr}
\bibliography{VAE_EVT}

\end{document}